\documentclass[conference]{IEEEtran}

\usepackage{graphicx}
\usepackage{amsmath}

\usepackage{caption}

\usepackage{algorithmic}
\usepackage{algorithm}

\usepackage{lipsum}
\usepackage[pscoord]{eso-pic}

\newcommand{\placetextbox}[3]{
	\setbox0=\hbox{#3}
	\AddToShipoutPictureFG*{
		\put(\LenToUnit{#1\paperwidth},\LenToUnit{#2\paperheight}){\vtop{{\null}\makebox[0pt][c]{#3}}}%
	}%
}%

\begin{document}
\title{A Machine learning approach for Shape From Shading }

\placetextbox{0.65}{1}{2nd International Conference on Signal, Image, Vision and their Applications (SIVA’13)}
\placetextbox{0.8}{0.985}{November 18-20, 2013 - Guelma, Algeria.}%

\author{\IEEEauthorblockN{Lyes ABADA}
\IEEEauthorblockA{Laboratory of Artificial Intelligence (LRIA)\\
Computer science department, USTHB, Algiers,Algeria\\
Email: labada@usthb.dz}
\and
\IEEEauthorblockN{Saliha AOUAT}
\IEEEauthorblockA{Laboratory of Artificial Intelligence (LRIA)\\
Computer science department, USTHB, Algiers,Algeria\\
Email: saouat@usthb.dz}
}
\maketitle

\begin{abstract}
The aim of Shape From Shading (SFS) problem is to reconstruct the relief of an object from a single gray level image. In this paper we present a new method to solve the problem of SFS using Machine learning method. Our approach belongs to Local resolution category. The orientation of each part of the object is represented by the perpendicular vector to the surface (Normal Vector), this vector is defined by two angles SLANT and TILT, such as the TILT is the angle between the normal vector and Z-axis, and the SLANT is the angle between the the X-axis and the projection of the normal to the plane. The TILT can be determined from the gray level, the unknown is the SLANT. To calculate the normal of each part of the surface (pixel) a supervised Machine learning method has been proposed. This method divided into three steps:  the first step is the preparation of the training data from 3D mathematical functions and synthetic objects. The second step is the creation of database of examples from 3D objects (off-line process). The third step is the application of test images (on-line process). The idea is to find for each pixel of the test image the most similar element in the examples database using a similarity value.
\end{abstract}


\section{Introduction}
Several researchers are working to solve the Shape From Shading(SFS) problem \cite{a_ma,a_Kunsberg,a_fan}, but there are no solutions that give good results on real images. Even with complex synthetic images, several constraints must be defined\cite{a_prados1}.

The methods of SFS can be classified into three categories: The first category concerns the local resolution methods (pontland\cite{a_pent1}\cite{a_denis1}, lee and Ronselfd\cite{a_denis1}, tsai and shah\cite{a_denis1}), in which, the computing of the surface orientation of each pixel is principally given by the gray level information of its neighbors. The second category is the global resolution methods in which the resolution is calculated using all the pixels of the image\cite{a_prados1,a_prados3}, by passing over each pixel several times. the third category is Mixed methods.

In this paper we are interested to the needle map integration methods \cite{sethian,a_denis4}, in which the resolution is given by two steps: the generation and then the integration of the needle map for the surface reconstruction (each step can be local global or mixed.

the needle map represents the set of normals corresponding to all the pixels of image. We propose a method for generating the needle map by using a Machine learning method. This method is composed of three phases: The first phase is the generation of the 3D object. The second phase is the preparation of the database examples (offline).In the third phase we use the database examples to generate the needle map of each pixel (online), it is belong to the local resolution methods.

Among the methods dealing with Shape from shading, we mention the Puntlands method \cite{a_denis1,a_pent1}, he proposed the first method to solve SFS problem using a local method. He chooses to direct all points of the image using the angles SLANT and TILT. There is also the method of lee and Ronselfd that follows the Puntlands method \cite{a_lee1,a_denis1,lee_zang}, but using a perspective camera and light source at the infinity. A.sethian\cite{sethian} is the first who applied the level set method to solve the SFS problem. His method uses the depth function (Z(x,y)) to generate the levels curves, and also jean Denis \cite{a_denis2} used this category where he suggested a perspective camera and light source at the infinity.

Like most methods proposed in SFS problem, we suppose that: the surface is smooth, the image taken by a parallel projection camera, light source at the infinity and regular surface (lambertian).

The paper is structured as follows: After the introduction section that offers a range of methods for solving the SFS problem with their classification. Section two summarizes the basic concepts for the image formation and explains the mathematical equations used. The third section details the proposed technique based on the Machine learning method. The last experimenting section gives the obtained results after applying our approach on synthetic and real images.

\section{Basic notions of image formation}
SFS is the process of generating a three-dimensional shape using a single two-dimensional image, which is the reverse process of the image formation.

\subsection{notations and definitions}
In this section we will define some basic notion and notations (see fig.\ref{fig:def}):

\begin{itemize}
\item Normal vector $N(N_{x},N_{y},N_{z})$: Is the perpendicular vector to the surface in a point (x,y).
\item light source vector $S(S_{x},S_{y},S_{z})$: Is the vector which represents the direction of the light source and its intensity, the direction of S is towards the light source.
\item needle map: Represents the set of the normals corresponding to all the pixels of image.
\item SLANT ($\theta$): is the angle between the origin (the X-axis) and the projection of the normal on the plane (x,y). 
\item TILT ($\phi$): Is the angle between the normal vector and Z-axis (see figure \ref{fig:def}).
\item Boundary Condition: The characteristic (solution) of the pixels located at the boundaries of the object is known.
\item Neumann boundary condition: The solution in Neumann Boundary Condition is the gradient (in our case the gradient of the depth Z).
\item Albedo: Reflectance factor (ratio of light emitted and light received).
\item Lambert surface: Surface that reflects radiation uniformly in all directions.
\item Gradient of depth (Z):depth variation (the derivative of Z with respect x and y).
\item Singular point: pixel has a maximum illuminance.

\end{itemize}

\begin{center}
\includegraphics[scale=0.55]{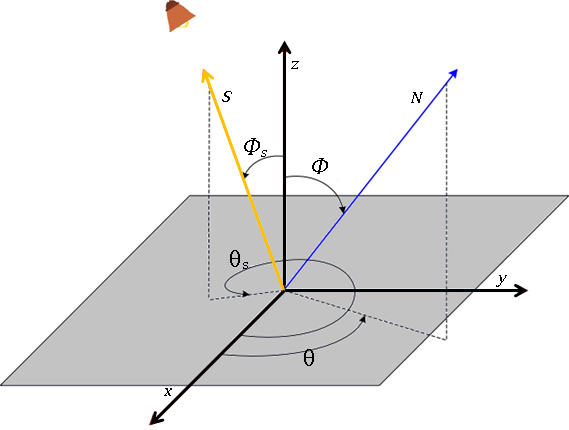}
\captionof{figure}{TILT and SLANT of normal and light source vectors}
\label{fig:def}
\end{center}

\subsection{Image formation}
The image formation consist to study the generation of images from objects (see fig.\ref{fig:defmodel}). This process is used in cameras. We can generate an image form a 3D object by using the basic equation of the images formation for more details see \cite{a_prados1}. The illumination E is:
\begin{eqnarray}
E=\frac{\alpha}{4}(\dfrac{p}{f})^{2}I\cos ^{4}\alpha L \label{eq:eq_1}
\end{eqnarray}

\begin{center}
\includegraphics[scale=0.5]{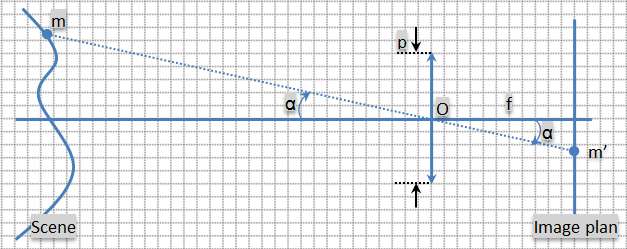}
\captionof{figure}{Perspective model of the camera}
\label{fig:defmodel}
\end{center}

The luminance (brightness) of a Lambert surface can be expressed as follows:

\begin{eqnarray}
L=\dfrac{\rho}{\pi}(\overrightarrow{N}.\overrightarrow{S})\label{eq:eq_5}
\end{eqnarray}




The illumination (E) can expressed using the angle $(\alpha)$ between the two vectors(N and S) \cite{sethian}:

\begin{eqnarray}
E=K(cos(\alpha)) \label{eq:eq_8}
\end{eqnarray}

K is a constant $(K =\dfrac{\rho}{4}(\dfrac{p}{f})^{2}I)$, $\rho$ is the albedo of the surface. In a singular point both vector N and S are equal, $(\alpha = 0)$, the illumination in this point is maximum $E_{max}=K$. The equation (~\ref{eq:eq_8}) can be written as \cite{sethian}:

\begin{eqnarray}
cos(\alpha)=\dfrac{E}{E_{max}} \label{eq:eq_10}
\end{eqnarray}

\subsection{Problem modeling}
Our method belongs to the category of integration methods which consists of reconstructing the scene in two steps. Generate the needle map then construct the scene from it. The following diagram in figure ~\ref{fig:schema1} shows the different Steps for the reconstruction.

\begin{center}
\includegraphics[scale=0.8]{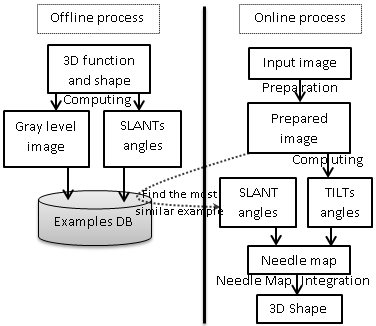}
\captionof{figure}{Steps for the 3D reconstruction}
\label{fig:schema1}
\end{center}

\subsection{Relation between normal vector, SLANT and TILT angles}

In our approach, we propose the generation of the needle map using the SLANT and TILT angles. We can compute the normal vector from the two angles (TILT and SLANT) using the formula: $N=(sin(\phi)cos(\theta),sin(\phi)sin(\theta),cos(\phi))$

As explained previously, the TILT angle is equal to the angle between the normal N and the light source. S, $TILT = \alpha = arccos(\dfrac{E}{E_{max}})$. We propose a method to compute the SLANT (the angle between the project vector of N on image plane and the vector N) using Machine learning under some constraints. We assume that the surface is Lambertian, differentiable, continuous with a punctual light source located at infinity, taken by a camera with a parallel projection.

\section{3D reconstruction using A Machine learning approach}

\subsection{Machine learning approach}
Our method is divided into three phases: the first phase is the generation of the 3D shape from mathematic functions. The second phase is the preparation of the examples Database (offline). Third phase uses the database examples to generate the needle map of the test images (online).

\subsubsection{Generation of 3D objects for learning}
In this phase we will generate the training data, these data contain the inputs (gray levels) and outputs (SLANT). In order to generate the data we need the 3D objects and their gray level images.

In this work we will create 3D objects by two methods: The first method is based on the generation of 3D surface from mathematic functions. Figures \ref{fig:app1}, \ref{fig:app2} and \ref{fig:app4} show three 3D objects generated by the functions $f1(x,y)=-x^{2}-y^{2}, f2(x,y)=x* exp(x^{2}-y^{2})$ and $f3(x,y)=sin(x)+sin(y)$ respectively. The second method uses 3D surfaces defined by their depth (Z) for example the silt (Figure.\ref{fig:app5}) Mozart (Figure\ref{fig:app6}) and penny (Figure\ref{fig:app7}). After the generation of the depths (Z of each pixels) we will calculate the TILT and SLANT angles, and then we will generate the image corresponding to each 3D object. We will finally get results as a set of pixels and corresponding angles.

\subsubsection{Offline phase}
The purpose of the offline phase is to create a database containing several examples. Each example contains:

\begin{itemize}
\item Input: 
	\begin{itemize}
		\item The gray level of the pixel (i,j)
		\item The gray level and the SLANT of three adjacent neighbors of pixel (i,j)
	\end{itemize}		
\item Output: The SLANT of pixel (i,j)
\end{itemize}
For example in figure \ref{fig:db}, we have a pixel (i,j) and three adjacent neighbors, The database contains eight fields, seven input and one output, the inputs are the Gray level of$\{(i,j);(i,j+1);(i+1,j);(i+1,j+1)\}$ and the SLANT of$\{(i,j+1);(i+1,j);(i+1,j+1)\}$. The SLANT of(i,j) is the output.

\begin{center}
\includegraphics[scale=0.8]{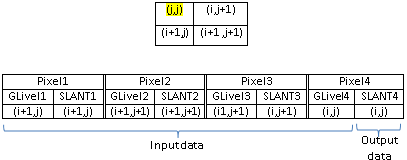}
\captionof{figure}{Database example}
\label{fig:db}
\end{center}

\subsubsection{Online phase}
In our approach we will use the Neumann boundary condition, so it is assumed that the normal of edge around the treated area in ​​the image is known (figure \ref{fig:bc}). All pixels with three adjacent neighbors known SLANT are ready to find its SLANT. The search is done by the Euclidean distance between the ready pixels and each element of examples database, and choose the minimum distances. 

{\footnotesize  $D(P(i,j),E(i,j)) = \\ \sqrt{ \sum \limits_{k=1}^3 (GLnP(k)-(GLnE(k))^{2}+(SnP(k)-(SnE(k))^{2}}$}

such as:
\begin{itemize}
\item $P(i,j)$: is a ready Pixel.
\item $E(i,j)$: is a database example .
\item $GLnP(k)$: is the gray level of $k^{th}$ neighbor of P(i,j)
\item $GLnE(k)$: is the gray level of $k^{th}$ neighbor of E(i,j)
\item $SnP(k)$: is the SLANT of $k^{th}$ neighbor of P(i,j)
\item $SnE(k)$: is the SLANT of $k^{th}$ neighbor of E(i,j)
\item $GP$ is the gray level of P(i,j)

\end{itemize}

\begin{center}
\includegraphics[scale=0.5]{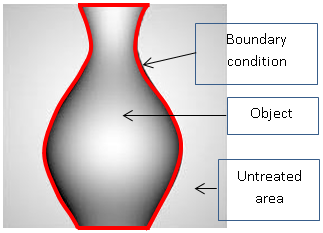}
\captionof{figure}{Boundary condition}
\label{fig:bc}
\end{center}

\subsection{Integration of the needle map for the scene reconstruction}
There are several methods to integrate the normal field, J.Denis\cite{a_denis4,a_denis5} shows some iterative and no-iterative methods of the normal field integration. Iterative methods are slow but give better results.
We use in the following the method of Horn and Brooks\cite{a_denis4}, it is simple, easy to implement and gives a good results. The equation of integration is follows:\\
$Z^{k}_{i,j}= \frac{Z^{k}_{i+1,j}+Z^{k}_{i,j+1}+Z^{k}_{i-1,j}+Z^{k}_{i,j-1}}{4}-\frac{\delta}{8g}(p_{i+1,j}-p_{i-1,j}+q_{i,j+1}-q_{i,j-1})$

p and q are calculated using SLANT($\theta$) and TILT($\phi$).

$(p,q)=(\dfrac{sin(\phi)cos(\theta)}{cos(\phi)},\dfrac{sin(\phi)sin(\theta)}{cos(\phi)})$

\section{Experiments}

The results of our approach is depend to the learning phase (offline), we will test on many examples database generated from different functions. Figures \ref{fig:app1}, \ref{fig:app2} and \ref{fig:app4} show four objects and corresponding images, these objects are generated using mathematics functions, $f1(x,y)=-x^{2}-y^{2}, f2(x,y)=x* exp(x^{2}-y^{2})$ and $f3(x,y)=sin(x)+sin(y)$. Figures \ref{fig:app5}, \ref{fig:app6} and \ref{fig:app7} shows three objects generated from the depths matrix and corresponding images. In all this cases the images are generated from objects using Equation \ref{eq:eq_8}.

\begin{center}
\includegraphics[scale=0.5]{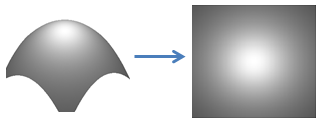}
\captionof{figure}{$f1(x,y)=-x^{2}-y^{2} \ / \{x,y\}\in [-1,1]$}
\label{fig:app1}

\includegraphics[scale=0.5]{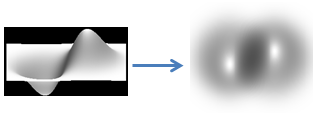}
\captionof{figure}{$f2(x,y)=x* exp(x^{2}-y^{2}) \ / \{x,y\}\in [-2,2]$}
\label{fig:app2}

\includegraphics[scale=0.5]{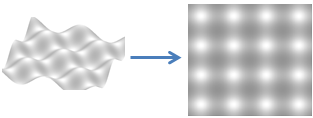}
\captionof{figure}{$f3(x,y)=sin(x)+sin(y) \ / \{x,y\}\in [-6,6]$}
\label{fig:app4}
\end{center}

\begin{center}
\includegraphics[scale=0.5]{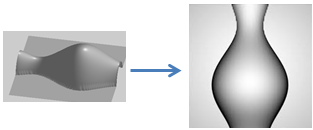}
\captionof{figure}{Depth matrix of silt}
\label{fig:app5}

\includegraphics[scale=0.5]{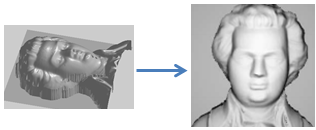}
\captionof{figure}{Depth matrix of mozart}
\label{fig:app6}

\includegraphics[scale=0.5]{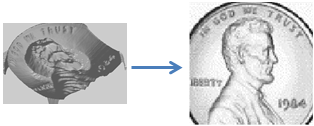}
\captionof{figure}{Depth matrix of penny}
\label{fig:app7}
\end{center}

We study six different cases of database, for each case we apply different function for the off-line and the on-line process:

\subsubsection{off-line: the function f1, on-line the image generated by f2}
First we will generate examples of function f1, knowing that the examples are disjoint, function f1 can be generate 4625 examples. The test result on the image generated by the function f2 is shown in Figure \ref{fig:res1}. The average of the distance of all pixels equals 0.07

\begin{center}
\includegraphics[scale=0.4]{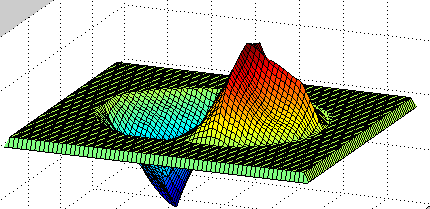}
\captionof{figure}{Result of image generating by f2}
\label{fig:res1}
\end{center}

\subsubsection{off-line: the functions:f1 and f2, on-line the image generated by f3}
Now we use the functions f1 and f2 to generate the examples database and f3 for the test. f1 and f2 generate 9249 example. The result is show in figure \ref{fig:res2} the average of distance is 0.064. So there are pixels that do not have examples nearest.

\begin{center}
\includegraphics[scale=0.4]{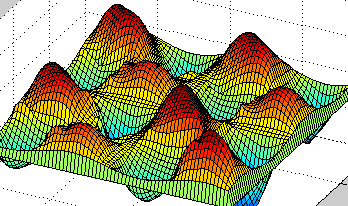}
\captionof{figure}{Result of image generated by f3}
\label{fig:res2}
\end{center}

\subsubsection{off-line: the functions f1,f2 and f3, on-line: the image of the silt}
In this case we will use f1, f2 and f3 in the database and the silt image \ref{fig:app5} for the test, the result is shown in Figure \ref{fig:res3}. The value of the average of distance is 0.1. Note that whenever the image is becoming more complicated the average distance increases.

\begin{center}
\includegraphics[scale=0.4]{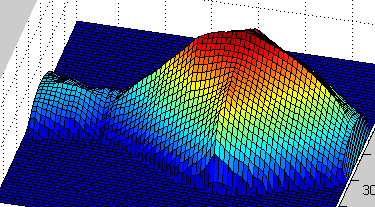}
\captionof{figure}{Result of silt image}
\label{fig:res3}
\end{center}

\subsubsection{off-line: the functions f1,f2 f3 and silt, on-line: the images mozart and penny:}
The Mozart and penny images contain more detail than the others, so there are several pixels that do not have a nearest example. Figure \ref{fig:res4} shows the test result on the two images. The average distance is higher 0.19 for the image of Mozart and 0.21 for the image of penny. The distance interval is between 0 and 1, it represents the difference between the gray level and the azimuth of a pixel in the test image and the nearest example in database. The distance is equal is to 1 if the difference equals $\pi or -\pi$. the distance of penny's image equal 0.21 and equal also 0.66 rad $(36^{\circ})$.


\begin{center}
\includegraphics[scale=0.4]{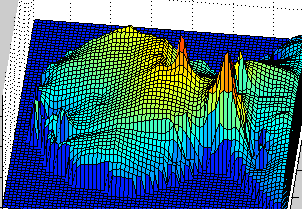}
\includegraphics[scale=0.35]{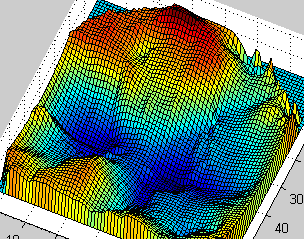}
\captionof{figure}{Result of Mozart and Penny images}
\label{fig:res4}
\end{center}

\subsubsection{off-line: all functions, on-line Silt,Mozart and penny}
The advantage of our approach is that we can add the examples to the database according to the use case. The result is depend on the the examples in the database. Now we take the best case, a database contain all possible examples, we will put all functions (f1,f2,f3,vase,mozart and penny) in the database, They generate 23004 examples. We test to the three images vase, Mozart and Penny, the average of the distance equal (0.0000554, 0.0004, 0.027), the results are shown in Figures \ref{fig:res21} \ref{fig:res22} \ref{fig:res23}.

\begin{center}
\includegraphics[scale=0.37]{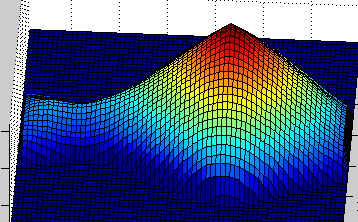}
\captionof{figure}{Result of image generated by the silt}
\label{fig:res21}
\end{center}

\begin{center}
\includegraphics[scale=0.37]{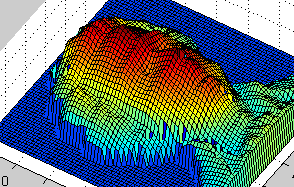}
\captionof{figure}{Result of image generated by Mozart object}
\label{fig:res22}
\end{center}

\begin{center}
\includegraphics[scale=0.37]{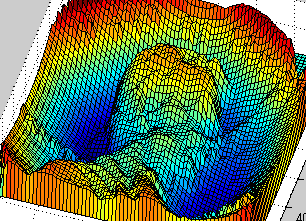}
\captionof{figure}{Result of image generated by penny object}
\label{fig:res23}
\end{center}

\subsubsection{use simple real images with database of all functions:} 
In the above results we tested our approach on synthetic images, in which the boundary conditions are known. To test our approach on images without boundary condition, we will use the edge of the object (there are several methods for detecting the edge). We assume that the projection is perpendicular to the tangent toward the outside. The results shown in Figures \ref{fig:res31} and \ref{fig:res32} are obtained with any additional information.

\begin{center}
\includegraphics[scale=0.7]{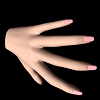}
\includegraphics[scale=0.35]{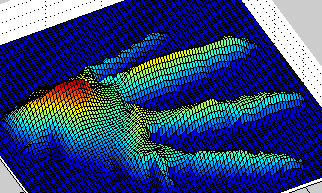}
\captionof{figure}{Result of image without Boundary Condition}
\label{fig:res31}
\end{center}

\begin{center}
\includegraphics[scale=0.8]{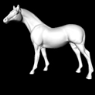}
\includegraphics[scale=0.35]{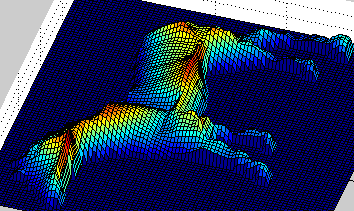}
\captionof{figure}{Result of image without Boundary Condition}
\label{fig:res32}
\end{center}




\section{Conclusion}

Most of the local resolution methods does not give a good results, because it is difficult to determine the depth variation from the gray level variation. The solutions of the local resolution are generally complex. In this work we proposed a simple local resolution method using Machine learning. It gives very acceptable results compared with other local resolution methods. The advantage of our approach is in the "learning phase". the examples database can be specialized (using same object types), i.e. we can create a database according to the use case.
In the future work we will improve this approach, so that we can test it on more complex images, and minimize the number of constraints.



\end{document}